# Comparative Analysis of Machine Learning Algorithms for Solar Irradiance Forecasting in Smart Grids

Saman Soleymani
Department of Electrical Engineering
Semnan University
Semnan, Iran
Saman_soleimany@semnan.ac.ir

Shima Mohammadzadeh
Department of Electrical Engineering
Islamic Azad University
Mashhad, Iran
shimaamohammadzadeh@gmail.com

*Abstract*— **The increasing global demand for clean and environmentally friendly energy resources has caused increased interest in harnessing solar power through photovoltaic (PV) systems for smart grids and homes. However, the inherent unpredictability of PV generation poses problems associated with smart grid planning and management, energy trading and market participation, demand response, reliability, etc. Therefore, solar irradiance forecasting is essential for optimizing PV system utilization. This study proposes the next-generation machine learning algorithms such as random forests, Extreme Gradient Boosting (XGBoost), Light Gradient Boosted Machine (lightGBM) ensemble, CatBoost, and Multilayer Perceptron Artificial Neural Networks (MLP-ANNs) to forecast solar irradiance. Besides, Bayesian optimization is applied to hyperparameter tuning. Unlike tree-based ensemble algorithms that select the features intrinsically, MLP-ANN needs feature selection as a separate step. The simulation results indicate that the performance of the MLP-ANNs improves when feature selection is applied. Besides, the random forest outperforms the other learning algorithms.**

*Index Terms*-- **Smart grids, solar irradiance forecasting, machine learning, feature selection.**

## I. INTRODUCTION

Solar energy has become a key source of sustainable electricity generation due to the growing need for renewable energy sources and concern over climate change [1]. Although solar energy is a clean and accessible resource, its stochastic nature and dependence on weather present distinct challenges to smart grids [2]. Solar radiation, which measures sun intensity reaching the surface of the Earth, plays an important role in determining solar energy production system effectiveness and predictability [3]. Solar irradiance forecasting has become a critical process in the smart grid sector [4]. This forecasting is essential for smart grid integration and stability [5], and reduces the negative impact of solar power instability on the grid, power electronic equipment [6], optimizing energy markets [7], energy storage management [8], transformers [9], and environmental concerns [10]. So, a comprehensive understanding of solar irradiance is essential to effectively integrating solar power into smart grids. It is essential to develop advanced forecasting methods that can provide reliable insights into future irradiance patterns in order for this understanding to be achieved.

In recent years, machine learning has emerged as a powerful tool for addressing the various and complex problems in smart grids, such as environment explosiveness level evaluation [11], smart monitoring systems [12], diagnosing location [13], generation [14], demand-side cooperation [15], and occupant estimation [16]. One of the most important applications of machine learning algorithms is their ability to forecast. A machine learning model can be trained using historical solar irradiance data, such as sophisticated meteorological variables and sensor-based data. The authors assess the use of hybrid machine-learning models for solar radiation forecasting [17]. To build a more precise forecasting model, they mix various machine learning techniques. Convolutional neural networks are utilized in [18] to forecast solar radiation by taking advantage of the spatial and temporal trends in the inputs and outputs to make better forecasts. An MLP-ANN algorithm is developed in [19] for solar irradiance forecasting based on meteorological data. In this work, feature importance is used to specify the most relevant features. Long Short-Term Memory is one of the various types of recurrent neural networks for sequential data, which are used in [20] for solar irradiance forecasting. [21] develops a support vector machine regression model to forecast solar irradiance.

Although significant studies have been conducted by using machine learning to predict solar irradiance, there are still significant aspects such as feature importance analysis, hyperparameter tuning, model validation, and testing on different unseen datasets that are crucial for developing accurate and robust forecasting models. Addressing these aspects can lead to more reliable predictions in solar irradiance forecasting that can improve the performance of the smart grids. The main contribution of this study can be summarized as follows:

- Advanced machine learning comparison: this study contributes by employing next-generation machine learning algorithms, including Random forest, Extreme Gradient Boosting (XGBoost), Light Gradient Boosted Machine (LightGBM) ensemble, Multilayer Perceptron Artificial Neural Networks (MLP-ANNs) and CatBoost. By utilizing these next-generation techniques, the research not only explores their individual capabilities but also provides a comprehensive comparison of their performance.

- Enhanced forecasting through feature selection: another key contribution is the incorporation of feature selection techniques to improve forecasting results.

- Hyperparameter tuning via Bayesian optimization: the research advances the area by using Bayesian optimization to adjust the hyperparameters of the machine learning models.

- Cross-Validation: this work further contributes by precisely testing the developed forecasting algorithms on diverse unseen datasets.

- Spatial generalization through multilocation testing: the final contribution is to test the developed models on different datasets collected from various geographical locations.

## II. METHODOLOGY

### A. Data collection and preprocessing

The data used to create the models used in this study comes from the National Solar Radiation Database [22]. The dataset contains 8 characteristics: solar irradiance, ambient temperature, pressure, humidity, wind speed wind direction, time, and length of day. Data are for 4 years from 2017 to 2021 at hourly intervals. After cleaning the datasets, it is a common practice to normalize the dataset. Normalization is a preprocessing technique that adjusts all the features to a similar scale, which can improve the performance and convergence of learning processes [23]. This study utilizes the standardization technique that transforms the data such that it has a mean of 0 and a standard deviation of 1 ensuring it follows the normal distribution [24].

### B. Feature selection

Feature selection is a technique in machine learning where a subset of the most relevant features or inputs is selected from a larger set of features. The goal of feature selection is to improve model performance, reduce computational complexity, enhance interpretability, and mitigate the risk of overfitting by focusing on the most relevant features [25]. The simplest way to remove the less significant and select the most relevant features is to train the machine learning through all possible combinations of features and choose the best one. However, this is an extremely time-consuming procedure. Accordingly, to select the most important features of a learning model, the feature selection technique is used. It should be noted that machine learning models like random forest, XGBoost, LightGBM, and CatBoost are based on the inherent feature importance provided by the algorithms during their training process [26]. That is, there is no need for a separate step to select the features. On the other hand, MLP-ANNs typically do not have the feature selection intrinsically. Feature selection techniques involve selecting features before feeding the data into the machine learning algorithms [27]. The selected features are then used for training. Among the most common feature selection techniques are wrapper and filter-based models. Many models are developed with all possible combinations of features, and the best model is selected based on a performance metric among all the models created by the wrapper feature selection models [28]. The present wrapper algorithms are forward, backward elimination, and recursive feature selection. Filter methods rely on statistical measures to rank and select features before applying a machine learning algorithm. Unlike wrapper methods that involve training and evaluating models, filter methods pre-process features independently of the chosen machine learning algorithm. Common filter methods include Pearson Correlation Coefficient (PCC), Chi-Squared test [29], and ANOVA analysis. This study utilized the PCC to select the most important features in order to feed them into the models. PCC is the ratio between the covariance of two variables and the product of their standard deviations and measures the linear correlation between a feature and the output variable. Features with high correlation coefficients are considered more relevant. The PCC value ranges between -1 and +1. A value of -1 indicates a perfect negative correlation, 0 indicates no correlation, and 1 indicates a perfect positive correlation [19].

As mentioned tree-based algorithms like random forest, XGBoost, LightGBM, and CatBoost typically do not need feature selection. However, the contribution of each feature on the final fitted model can be demonstrated. To do that, Shapley's Additive explanation (SHAP) plot analysis is performed. SHAP plot is utilized to visualize the feature importance in the random forest, XGBoost, LightGBM, and CatBoost. This approach enhances the ability to interpret and validate the model decision-making process.

### C. Learning algorithms

Learning algorithms are powerful tools that can be used for forecasting due to their learning ability. Random forest, MLP-ANNs, LightGBM, XGBoost, and CatBoost are utilized in this study. Random forest is an ensemble algorithm that uses a collection of decision trees for classification and regression. It uses a method called bagging (bootstrap aggregation) to reduce variance while maintaining low bias. LightGBM is a fast, distributed, high-performance gradient boosting framework based on the decision tree algorithm, used for ranking, classification, and many other machine learning tasks. XGBoots is another powerful machine learning model that constructs an ensemble of decision trees sequentially, optimizing a loss function and handling data imbalance, missing values, and overfitting. MLP-ANNs are a type of deep learning model composed of multiple interconnected layers and neurons [30]. It is designed to capture complex patterns in data. The main idea of CatBoost is to sequentially combine many weak models and, thus, create a solid competitive predictive model through greedy search.

### D. Hyperparameter tuning

A practice to select the best hyperparameters in the learning models to improve performance is known as hyperparameter tuning. This study utilized Bayesian optimization which has been able to successfully find the optimal combination of hyperparameters of the learning models [29]. Bayesian optimization is a probabilistic model to predict which hyperparameters will lead to better performance and then tests those predictions. To implement Bayesian optimization in learning algorithms, a common practice involves utilizing the k-fold cross-validation technique. A piece of the training dataset is assigned for validation, and the model is trained iteratively over k-folds. During each iteration, one set is assigned for validation, while the remaining k-1 folds are used for training. Model performance is averaged across these k iterations to mitigate variability. In each iteration, Bayesian optimization seeks to determine hyperparameters that optimize the forecasting results.

### E. Evaluation metrics

Evaluation metrics are used to measure the quality of the machine learning models. This study uses Mean Absolute Error (MAE), Root Mean Square Error (RMSE), and coefficient of determination or $R^2$ [32] to assess the performance of the models.

### III. SIMULATION AND RESULTS

The data are split into three sets: training, cross-validation, and testing datasets. 80% of the data are selected as the training set and 20% as the test. in addition, cross-validation is performed on 20% of the training datasets. In this work, the input data are first analyzed by the PCC algorithm, and the data with the highest correlation with solar radiation are identified. This data is then used as input for the MLP-ANN model. Fig. 1 demonstrates the PCC results. Based on Fig. 1 variables such as ambient temperature, humidity, and wind direction show a substantial correlation with solar irradiance, indicating a higher impact on solar irradiance forecasting. While features like pressure and wind speed exhibited lower influence. Length and time of day could be considered redundant features. To assess the impact of the feature selection technique on the performance of the models, the MLP-ANNs are trained and tested with and without feature selection. The optimal hyperparameters of the fitted MLP-ANNs with PCC by utilizing Bayesian optimization are provided in Table I. Besides, Table II shows solar irradiance forecasting results with and without PCC feature selection. MLP-ANN model performance is enhanced by PCC-based feature selection as MAE, RMSE decreased and $R^2$ increased, indicating improved performance. Table II shows the results of predicting solar radiance using random forest, XGBoost, LightGBM, and CatBoost. The results given in Table II compare the performance of various algorithms for forecasting solar radiance. Random forest performs the lowest RMSE of 87.18 MAE of 34.82 and highest $R^2$ of 0.95, indicating superior predictive accuracy. LightGBM and XGBoost algorithms also demonstrated competitive performance, with RMSE values of 111.49 and 114.01, MAE values of 50.16 and 51.57, and $R^2$ values of 0.87 and 0.86, respectively. These results highlight the effectiveness of ensemble techniques like random forest, LightGBM, and XGBoost for accurate solar radiance prediction, while also emphasizing the significance of considering feature importance in the MLP-ANN model.

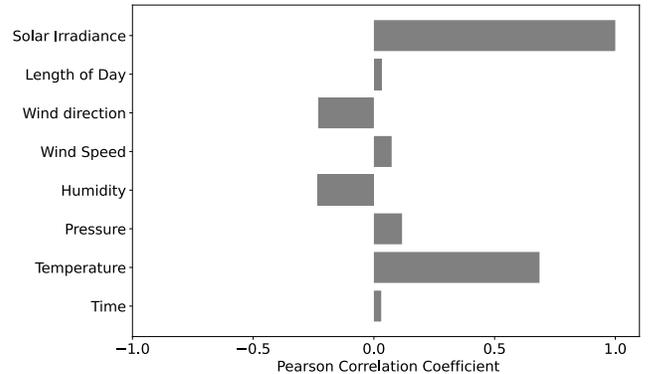

Figure. 1. PCC values calculated between solar irradiance and various meteorological parameters.

Because random forest produces favorable results, further investigation of its generalization capabilities is warranted. Through a five-fold cross-validation process, potential overfitting problems are addressed. As a result of this analysis, we can observe whether the observed performance is consistent for different training set sizes and their corresponding validation scores. Fig. 2 demonstrates the learning curve associated with the fitted random Forest algorithm. The training

TABLE I. Characterization of MLP-ANN parameters with PCC.

| hyperparameter | value |
|---|---|
| Solver | adam |
| alpha | 0.0001 |
| batch_size | Auto |
| learning_rate | 0.001 |
| power_t | 0.5 |
| max_iter | 5000 |
| shuffle | true |
| Random state | 42 |
| Hyperparameter optimizer | Bayesian |
| Loss function | Cross-Entropy |
| Activation function | Relu |
| Neurons of input layers | 15 |
| Neurons of the first hidden layers | 10 |
| Neurons of second hidden layers | 5 |
| Neurons of third hidden layers | 5 |
| Neurons of output layers | 1 |

TABLE II. Characterization of MLP-ANN parameters with PCC.

| Algorithm | RMSE | MAE | $R^2$ |
|---|---|---|---|
| MLP without PCC | 124.50 | 55.41 | 0.83 |
| MLP with PCC | 95.16 | 41.45 | 0.91 |
| Random Forest | 87.18 | 34.82 | 0.95 |
| LightGBM | 111.49 | 50.16 | 0.87 |
| XGBoost | 114.01 | 51.57 | 0.86 |
| CatBoost | 130.42 | 60.12 | 0.79 |

TABLE III. Characterization of the random forest parameters.

| hyperparameter | value |
|---|---|
| criterion | Gini |
| Random-state | 0 |
| n_estimators | 400 |
| min_impurity_decrease | 0 |
| max_depth | none |
| bootstrap | true |
| max_features | auto |

MAE values range from 5 to 32 as the training set size increases. Furthermore, the validation MAE values are decreasing from 230 to 34.2. There is a considerable gap between the training and validation MAE values at the beginning of the curve (when training sets are smaller), which may indicate overfitting, in which the model performs remarkably well on the training data but struggles to generalize to unseen data (higher validation MAE). However, as the training set size increases the validation MAE decreases and the gap narrows, indicating that the generalization ability of the model is improved. Additionally, as training and validation MAE values converge, it indicates improved generalization and reduced overfitting concerns. As a result, it is evident from the learning curve that the random forest algorithm provides accurate and consistent solar irradiance forecasts. Besides, the optimal hyperparameters of the fitted random forests by utilizing Bayesian optimization are provided in Table III.

Further, a SHAP plot is also created to assess the importance and contribution of each feature on the trained model in Fig 3. The SHAP plot demonstrates the distribution of each feature over all samples in the dataset. The light gray and gray areas represent features with low and high values, respectively. The Y-axis represents the importance of the features on solar irradiance forecasting. Besides, the features on the left X-axis are ordered based on their mean absolute SHAP values. These orders are consistent with the PCC results indicating that the random forest captures the relevant features. The ambient temperature has the highest SHAP value, indicating its high contribution to solar irradiance forecasting. In addition, the higher the ambient temperature values, the higher the impact on model output. Following the ambient temperature, the second and third significant features are wind direction and humidity which have a negative correlation to the target values. Features like wind speed and air pressure play intermediate roles, contributing to solar irradiance prediction. As the values of wind speed and air pressure increase (from blue to red), their corresponding SHAP values increase, indicating a positive correlation with target values. Finally, the time and length of the day have minimal impact on solar irradiance forecasting.

To further evaluate the robustness of the fitted random forest model, it is tested on different locations to assess its capability to generalize across altering environmental conditions and geographical areas. Fig. 4 shows the RMSE and MAE of the fitted random forest for the original dataset location and two different locations that are 100 and 500 miles away from the original location. The model performance decreases as the distance from the original location increases. Both RMSE and MAE are slightly increased by a 100-mile displacement. Further, a distance of 500 miles shows more increase in model performance parameters. The slight decrease in model performance can be attributed to varying geographical and climatic conditions. Despite the decreasing performance, random forests still outperform LightGBM, XGBoost, CatBoost, and MLP-ANN in this case. Random forest models are particularly robust to varying geographical and climate conditions because they are intrinsically adaptable to complicated relationships between input features and solar radiation.

IV. CONCLUSION

In this study, we explored the use of advanced machine learning algorithms for forecasting solar irradiance, a key factor for efficient smart grid management. We applied the PCC for feature selection and Bayesian optimization for fine tuning the learning algorithms. The findings reveal that the MLP-ANNs perform better when coupled with PCC-based feature selection. The random forest algorithm emerged as the most accurate in our tests, even in different geographical conditions. Future work could explore different feature selection methods, such as wrapper and filter methods, and other ensemble algorithms for potentially better accuracy. Additionally, alternate optimization techniques can be examined to improve the model's performance. Furthermore, using more parameters such as the temperature of the earth's crust, pollution index and cloud index can improve the forecasting.

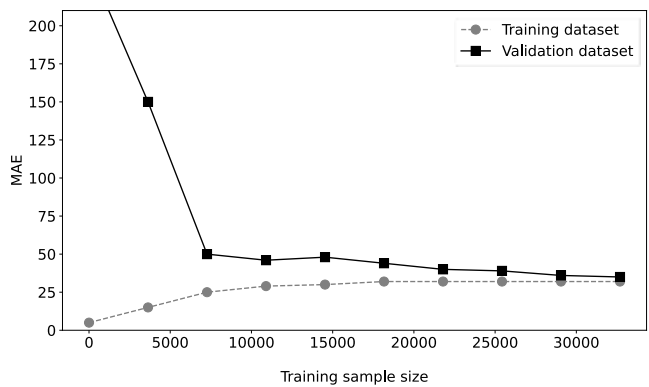

Figure. 2. Learning curve associated with the fitted random forest.

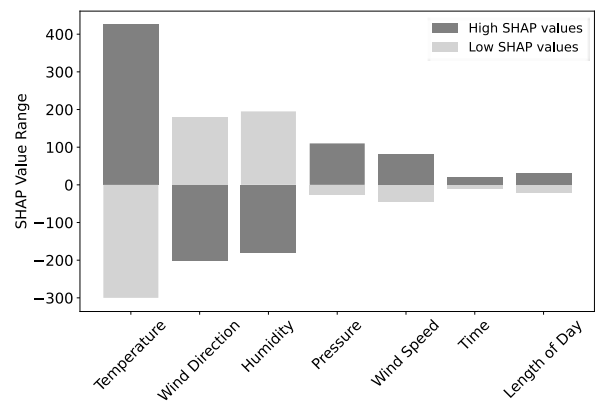

Figure. 3. SHAP plot associated with the fitted Random forest

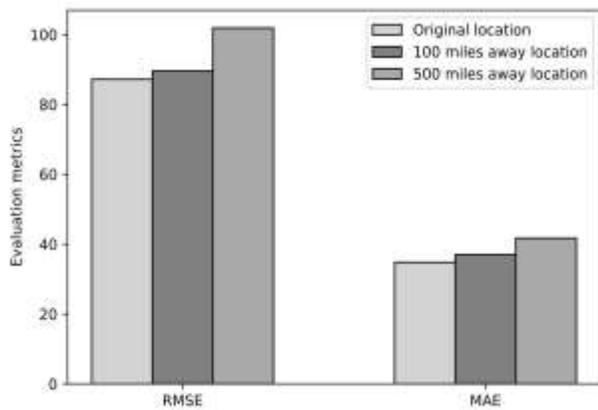

Figure. 4. RMSE and MAE of the random forest for different locations.